\documentclass[sigconf,natbib=true]{acmart}
\makeatletter
\def\@ACM@checkaffil{
    \if@ACM@instpresent\else
    \ClassWarningNoLine{\@classname}{No institution present for an affiliation}%
    \fi
    \if@ACM@citypresent\else
    \ClassWarningNoLine{\@classname}{No city present for an affiliation}%
    \fi
    \if@ACM@countrypresent\else
        \ClassWarningNoLine{\@classname}{No country present for an affiliation}%
    \fi
}
\makeatother

\usepackage[ruled,linesnumbered]{algorithm2e}
\usepackage{multirow}
\usepackage{enumitem}
\usepackage{makecell}
\usepackage{balance}
\usepackage{amsfonts}
\usepackage{pifont}

\AtBeginDocument{%
  \providecommand\BibTeX{{%
    \normalfont B\kern-0.5em{\scshape i\kern-0.25em b}\kern-0.8em\TeX}}}

\copyrightyear{2023}
\acmYear{2023}
\setcopyright{rightsretained}
\acmConference[MM '23]{Proceedings of the 31st ACM International Conference on Multimedia}{October 29-November 3, 2023}{Ottawa, ON, Canada}
\acmBooktitle{Proceedings of the 31st ACM International Conference on Multimedia (MM '23), October 29-November 3, 2023, Ottawa, ON, Canada}
\acmDOI{10.1145/3581783.3611899}
\acmISBN{979-8-4007-0108-5/23/10}

\makeatletter
\gdef\@copyrightpermission{
  \begin{minipage}{0.3\columnwidth}
   \href{https://creativecommons.org/licenses/by/4.0/}{\includegraphics[width=0.90\textwidth]{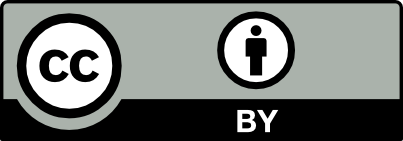}}
  \end{minipage}\hfill
  \begin{minipage}{0.7\columnwidth}
   \href{https://creativecommons.org/licenses/by/4.0/}{This work is licensed under a Creative Commons Attribution International 4.0 License.}
  \end{minipage}
  \vspace{5pt}
}
\makeatother
\begin{document}

\newcommand{\modelname}{\texttt{MR$^{\texttt{2}}$}}

\newcommand{\cmark}{\ding{51}}%
\newcommand{\xmark}{\ding{55}}%


\title{Prompt Me Up: Unleashing the Power of Alignments for Multimodal Entity and Relation Extraction}

\author{Xuming Hu}
\authornote{Equal contribution. Source code: \url{https://github.com/THU-BPM/PROMU}.}
\affiliation{%
  \institution{Tsinghua University}
}
\email{hxm19@mails.tsinghua.edu.cn}

\author{Junzhe Chen}
\authornotemark[1]
\affiliation{%
  \institution{Tsinghua University}
}
\email{chenjz20@mails.tsinghua.edu.cn}

\author{Aiwei Liu}
\affiliation{%
  \institution{Tsinghua University}
}
\email{liuaw20@mails.tsinghua.edu.cn}

\author{Shiao Meng}
\affiliation{%
  \institution{Tsinghua University}
}
\email{msa21@mails.tsinghua.edu.cn}

\author{Lijie Wen}
\authornote{Corresponding author.}
\affiliation{%
 \institution{Tsinghua University} %
 }
\email{wenlj@tsinghua.edu.cn}

\author{Philip S. Yu}
\affiliation{%
 \institution{University of Illinois at Chicago}
 }
\email{psyu@cs.uic.edu}

\renewcommand{\shortauthors}{Xuming Hu et al.}

\begin{abstract}
How can we better extract entities and relations from text? Using multimodal extraction with images and text obtains more signals for entities and relations, and aligns them through graphs or hierarchical fusion, aiding in extraction. Despite attempts at various fusions, previous works have overlooked many unlabeled image-caption pairs, such as NewsCLIPing. This paper proposes innovative pre-training objectives for entity-object and relation-image alignment, extracting objects from images and aligning them with entity and relation prompts for soft pseudo-labels. These labels are used as self-supervised signals for pre-training, enhancing the ability to extract entities and relations. Experiments on three datasets show an average 3.41\% F1 improvement over prior SOTA. Additionally, our method is orthogonal to previous multimodal fusions, and using it on prior SOTA fusions further improves 5.47\% F1.
\end{abstract}





\maketitle

\section{Introduction}
\label{sec:intro}

\begin{figure}
    \centering
    \includegraphics[scale=0.21]{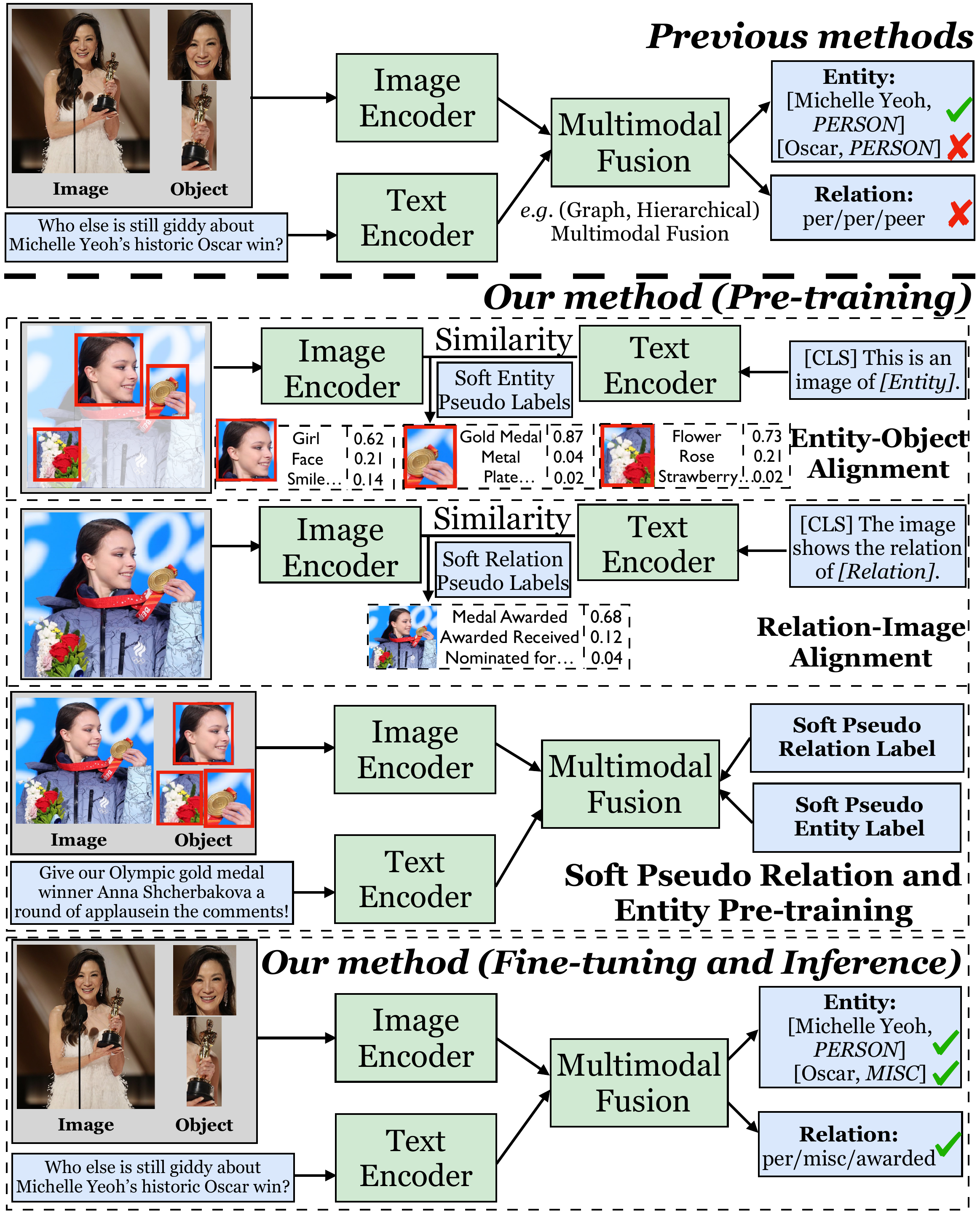}
    \vspace{-2mm}
    \caption{The previous methods (above) utilized graphs or hierarchical multi-modal fusion to combine text and image embeddings. Our method (below) aligns soft entity pseudo-label prompts by extracting potential objects and soft relation pseudo-label prompts through images. The obtained self-supervised pseudo-label signals are used to pre-train the multimodal fusion, enhancing its ability to extract entity and relation-related information from images.}
    \label{fig:example}
    \vspace{-5mm}
\end{figure}

Entity and relation extraction aim at detecting potential entities and their inter-relations from unstructured text, building structured \textit{<Entity, Relation, Entity>} triples for downstream tasks like QA \cite{song2018explore,liu2022semantic,liu2023comprehensive} and web search \cite{lokovc2019framework}. Prior attempts solely relied on single modality, e.g., text, for extraction \cite{hu2020selfore,hu2021semi,hu2021gradient,liu2022hierarchical,hu2023selflre}. Recent studies found that image representations aid triple extraction, prompting multimodal extraction research \cite{zheng2021mnre,chen2022good,lin2022multimodal,hu2023multimodal}. Fusion and alignment of image and text representations have become a research focus in the MM community. \citet{zhang2021multi} and \citet{zheng2021multimodal} attempted graph-based multimodal fusion for fine-grained object and entity information extraction. \citet{chen2022good} tried a hierarchical multimodal fusion framework, removing irrelevant image objects, enhancing robustness and effectiveness of entity extraction, achieving SOTA results. Overall, all works have been confined to training multimodal models based solely on existing datasets. However, due to the scarcity of labeled multimodal data, these works' effectiveness has been significantly constrained.

Addressing this, large-scale unsupervised image-text pairs like NewsCLIPping \cite{escalada2010newsclipping} serve as an effective solution for pretraining multimodal models. Multimodal pre-training works like CLIP \cite{radford2021learning}, Oscar \cite{li2020oscar}, and AlPro \cite{li2022align} offered solutions by aligning text with images, but neglect modeling entity and relation information embedded within modalities. We propose a multi-modal pre-training method focusing on entity and relational information within text and image, using a vast amount of unlabeled image-caption pairs and novel alignment methods to obtain pseudo entity and relation labels for entity-object and relation-image pairs.
These pairs help in aligning the semantic spaces of images and text and are beneficial for the extraction of multimodal entities and relations. As shown in Figure \ref{fig:example}, the entity-object alignment task uses automated object recognition tools, such as YOLO \cite{wang2022yolov7}, to extract potential objects from images. We then use automated entity detection tools, like spaCy (\url{https://spacy.io/}), to extract potential entities from captions and align these entities with potential objects using entity prompts, such as ``This is an image of [Entity]'', to obtain soft pseudo entity labels. For example, when the object is a girl, the ranked entity embedding similarities are: [Girl, 0.62; Face, 0.21; ...]. We store the pseudo entity labels corresponding to the object. Similarly, the relation-image alignment task collects a relation database from Wikidata and calculates similarity between relation prompts, such as ``The image shows the relation of [Relation]'', and image embeddings. For instance, for the image in Figure \ref{fig:example}, we can obtain pseudo relation labels: [Medal Awarded, 0.68; Awarded Received, 0.12; ...].

After obtaining these pseudo entity and relation labels, we can use them as self-supervised signals to guide the pretraining of multimodal fusion. The soft label mechanism smooths the impact of prediction errors, even if some pseudo labels are incorrect \cite{muller2019does}. We employ cross-entropy loss to enhance the multimodal fusion's ability to extract objects and their relations in images and guide the text representations of entities and their relations. Take Figure \ref{fig:example} as an example, previous methods incorrectly predicted the entity \textit{Oscar} as the \textit{PERSON} type, leading to an erroneous relation prediction of \textit{per/per/peer}. In our approach, the multimodal fusion has already unleashed the power of alignment information between potential objects and entities from a massive amount of unlabeled image-caption pairs, so it easily predicts \textit{Oscar} as the \textit{MISC} type, resulting in the correct relation prediction of \textit{per/misc/awarded}.

Our experiments on three public multimodal NER and RE datasets showed a 3.41\% F1 score improvement over previous SOTA methods. Our self-supervised pretraining is orthogonal to prior hierarchical and graph-based multimodal fusion techniques, using vast image-caption data to pretrain SOTA fusion modules. Continued pretraining enhances F1 performance by 5.47\% for NER and RE tasks. We've experimented with Visual ChatGPT \cite{wu2023visual} in multimodal NER and RE tasks due to the explosive popularity of ChatGPT. In summary, our contributions are threefold: 


\vspace{-2mm}
\begin{itemize}
   \item We innovatively propose entity-object and relation-image alignment pretraining tasks, enabling the extraction of self-supervised signals from massive unlabeled image-caption pairs to pretrain multimodal fusion modules and improve NER and RE performance. 
   \item We obtain soft pseudo entity and relation labels through entity and relation prompting methods, which smooth out the impact of prediction errors and provide high-quality self-supervised signals.
   \item Our experiments show that our pretraining method can achieve an average F1 improvement of 3.41\% compared to previous SOTA methods, and our pretraining approach is orthogonal to earlier SOTA techniques. Using our pretraining method on previous SOTA approaches can further boost F1 scores by 5.47\%.
\end{itemize}

\section{Related Work}
\label{related}

\subsection{Multimodal Entity and Relation Extraction} 
As crucial components in information extraction, Named Entity Recognition (NER) and Relation Extraction (RE) have garnered significant interest in research. Prior studies mostly focused on textual modality, extracting entities and relations from single text sources \cite{miranda2020named,hu2023entity,hu2023gda,hu2023think}. However, given the impressive capabilities of multimodal data in numerous fields, studies concentrating solely on text have become limited \cite{chen2022good,zheng2021multimodal,zheng2021mnre}. Recently, some research has sought to integrate image modality with text for better entity extraction and relation identification, leading to multimodal NER and RE tasks. \citet{zhang2018adaptive,lu2018visual,moon2018multimodal,arshad2019aiding} proposed RNN-based text encoding, CNN-based image encoding, and implicit interactions to simulate information fusion between modalities, thereby enhancing entity extraction accuracy.

More recently, \citet{yu2020improving,zhang2021multi} proposed using region-based image features to represent objects, employing Transformer and visual encoders for fine-grained semantics. Numerous experiments showed that including objects from images can significantly improve entity extraction accuracy. However, most methods overlook error sensitivity. \citet{sun2021rpbert} proposed learning a text-image relation classifier for better multimodal fusion and reducing irrelevant image interference. \citet{chen2022good} attempted a hierarchical multimodal fusion framework to remove irrelevant objects. \citet{wang2022named} used a retrieval approach to gather relevant text from Wikipedia for image and text-based predictions.

Nonetheless, previous methods neglected the vast amount of easily accessible image-caption pair data, such as NewsCLIPing. In this paper, we innovatively propose entity-object and relation-image alignment tasks, extracting soft pseudo entity and relation labels from image-caption pair data as self-supervised signals to enhance the extraction capabilities of multimodal fusion models.

\begin{figure*}
    \centering
    \includegraphics[scale=0.48]{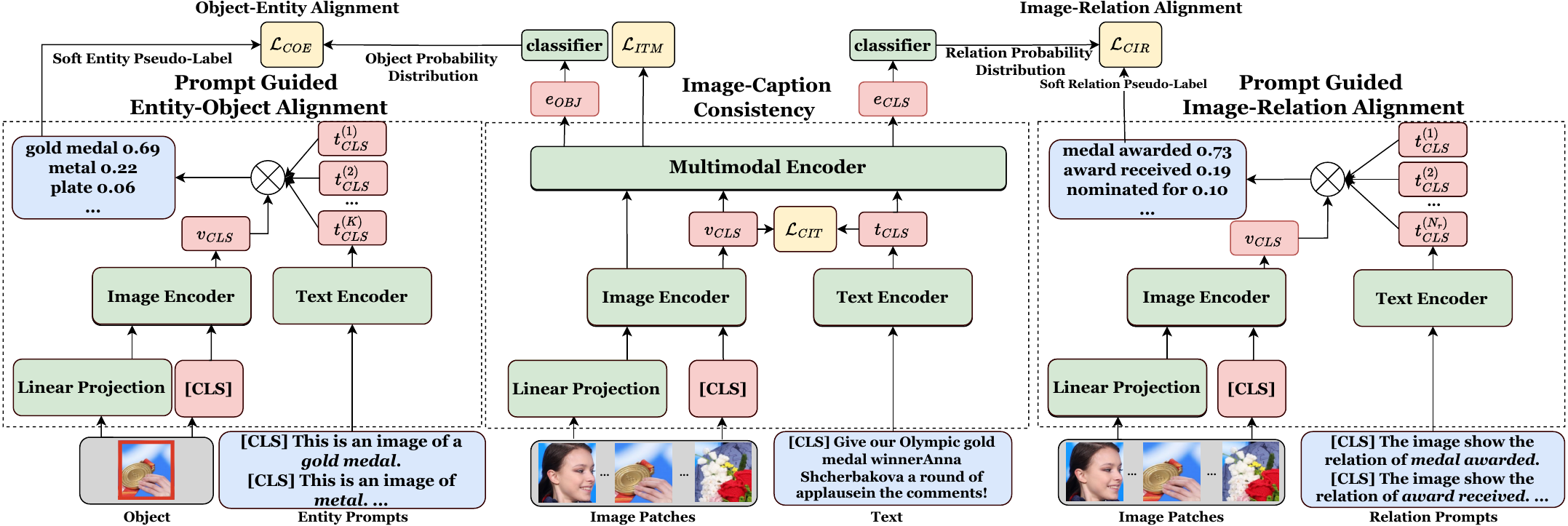}
    \vspace{-2mm}
    \caption{Overview of the multimodal pretraining. We utilize three contrastive losses: the contrastive image-text loss ($\mathcal{L}_{CIT}$), the contrastive object-entity loss ($\mathcal{L}_{COE}$), and the contrastive image-relation loss ($\mathcal{L}_{CIR}$). These losses aim to align the input text sequence with the corresponding image, align the entities and objects, and align the image with pre-defined relations respectively. We pre-trained the model on NewsCLIPping \cite{escalada2010newsclipping}, which includes both matched and mismatched image-caption pairs. In the matched pairs, the caption accurately describes the image, resulting in similar representations in semantic space, and vice versa. Therefore, we additionally utilize an image-text match loss ($\mathcal{L}_{ITM}$).}
    \label{fig:arch}
    \vspace{-4mm}
\end{figure*}

\vspace{-2mm}
\subsection{Self-Supervised Multimodal Learning} 
In multimodal learning, models process and integrate data from multiple modalities \cite{baltruvsaitis2018multimodal,ramachandram2017deep,bayoudh2021survey}, with applications in visual and language learning \cite{park2023visual}, video understanding \cite{schiappa2022self,selva2023video}, and natural language understanding \cite{li2022pair,li2023multi,liu2023exploring}. However, expensive human annotations are often required for effective training. Self-supervised learning \cite{alayrac2020self,thapa2022survey,schiappa2022self,zong2023self} has addressed this by using one modality as a supervisory signal for another, such as masking elements in images or text and using information from the other modality to reconstruct the masked content \cite{amrani2021noise,akbari2021vatt}.

Self-Supervised Multimodal Learning (SSML) leverages multimodal data and self-supervised objectives to enhance multimodal models. Pairings between modalities can be used as input by SSML algorithms (e.g., when one modality supervises another \cite{kim2021vilt,chen2020uniter}) or as output (e.g., learning from unpaired data and inducing pairing as a byproduct \cite{hu2019deep,wei2022inter}). Classical SSML methods include coarse-grained and fine-grained alignments; coarse-grained alignment assumes alignment between images and captions in multimodal self-supervision \cite{radford2021learning}; fine-grained alignment refers to correspondence between caption words and image patches \cite{li2020oscar,li2022align}. However, separate embedding spaces for modalities decrease effectiveness in modeling cross-modal interactions. In this paper, we employ entity and relation prompts for both alignments and use soft pseudo entity and relation labels to train multimodal fusion, improving entity and relation extraction performance.
\vspace{-2mm}

\section{Task Formulation}
\label{sec:task}
The multimodal named entity recognition (MNER) is defined as a sequence labeling task. Given an input text sequence $X=\{x_1,x_2,\cdots,x_{N}\}$ and an associated image $V$, where $N$ is the length of the sequence, the objective of the MNER task is to identify entities in the input sequence using the \texttt{B}, \texttt{I}, \texttt{O} format and categorize the recognized entities into predefined types $C=\{c_1,c_2,\cdots,c_{N_c}\}$. $N_c$ is the number of types and \texttt{B}, \texttt{I}, or \texttt{O}, indicates whether it is the beginning of an entity, inside an entity, or outside of any entity, respectively.  The output of the MNER task is a sequence of labels $Y=\{y_1,y_2,\cdots,y_N\}$, where $y_i \in \mathcal{Y}=\{B-c, I-c, O\}$ and $c\in C$, representing the label of the $i$-th input token. 

The multimodal relation extraction (MRE) task involves an input text sequence $X=\{x_1,x_2,\cdots,x_{N}\}$ with two pre-extracted named entities $e_1$ and $e_2$, as well as visual content $V$, where $N$ is the length of the text sequence. The objective of the MRE task is to classify the corresponding relation tag $r\in R=\{r_1,r_2,\cdots,r_{N_r}\}$ between $e_1$ and $e_2$. The set $R$ contains the pre-defined relation types, and $N_r$ represents the number of the relations.


\section{Model Architecture}
\label{sec:model}
We introduce two prompt-guided alignment mechanisms, one for entity-object alignment and the other for image-relation alignment, in addition to an image-caption consistency alignment module. These alignment modules are pre-trained on large-scale multimodal data and then fine-tuned on task-specific data for NER and RE. In the following sections, we describe each of these modules in detail.
\subsection{Image-Caption Consistency Alignment}
\label{section:pre-training model}
The goal of this module is to encode the input text and image in a way that their respective features are mapped to the same embedding space, allowing for improved fusion between the two modalities. Given a batch of matched and mismatched image-caption pairs obtained from NewsCLIPpings $D=\{(X_i, V_i, y_i)\}_{i=1}^{B}$, where B denotes the batch size, $X_i=\{x^{(i)}_1,x^{(i)}_2,\cdots,x^{(i)}_N\}$ is the caption that describes the image $V_i$ and $y_i$ is a binary label indicating whether the pair is matched (1) or not (0), we utilize a context encoder to obtain contextualized representations of the input caption. The context encoder can be a Transformer~\citep{vaswani2017attention} or a BERT~\citep{devlin2019bert}, where we use the BERT as an example. We feed the text sequence $X_i$ into BERT to obtain the encoded representation $\boldsymbol{t_i} = BERT(X_i)$, where $\boldsymbol{t_i} = \{t^{(i)}_{CLS}, t^{(i)}_1, t^{(i)}_2, \cdots, x^{(i)}_N\}\in\mathbb{R}^{d\times(N+1)}$ denotes the encoded representation of the input sequence with hidden embedding size $d=768$. Here, we use the representation of the special token \texttt{[CLS]} $t^{(i)}_{CLS}$ as the textual representation. 

To encode the visual content $V_i$, we first partition it into $K$ disjoint small patches $P_i=\{p^{(i)}_1,p^{(i)}_2,\cdots,p^{(i)}_K\}$ and apply a projection head $proj(\cdot)$ to map each patch into a lower-dimensional space to obtain a sequence of patch tokens $m_i$. Then, we use a pre-trained Vision Transformer (ViT)~\citep{DosovitskiyB0WZ21} trained on ImageNet ~\citep{DengDSLL009} to encode these tokens and generate a sequence of visual embeddings:
\begin{equation}
\begin{split}
\boldsymbol{m}_i = proj(P_i),\\
\boldsymbol{v}_i = Vit(\boldsymbol{m}_i),
\end{split}
\end{equation}
where $\boldsymbol{m}_i\in\mathbb{R}^{d_m\times K}$ and $\boldsymbol{v}_i=\{v^{(i)}_{CLS},v^{(i)}_1,v^{(i)}_2,\cdots,v^{(i)}_K\}\in\mathbb{R}^{d_v\times (K+1)
}$. Here, $d_m$ and $d_v$ represent the hidden size of the projection head and the Vision Transformer, respectively.
Similar to text encoding, we obtain the visual representation by using the output of the \texttt{[CLS]} token, denoted as $v^{(i)}_{CLS}$.

To obtain multimodal embeddings, we concatenate the text embeddings and visual embeddings previously obtained and then use a BERT model to fuse the embeddings from the two modalities:
\begin{equation}
\begin{split}
\boldsymbol{e}_i = BERT(concat(\boldsymbol{t}_i,\boldsymbol{v}_i)),
\end{split}
\end{equation}
where $\boldsymbol{e} = \{e^{(i)}_{CLS},e^{(i)}_1,e^{(i)}_2,\cdots,e^{(i)}_{K+N}\} \in \mathbb{R}^{d\times(N+K+1)}$ is the output multimodal embeddings and $d$ is the hidden shape of 768.

Firstly, we employ an image-text matching loss $\mathcal{L}_{ITM}$ to aid in the alignment between images and captions of the model:
\begin{equation}
\begin{split}
p_i &= fc(\boldsymbol{e}_i^{CLS}),\\
\mathcal{L}_{ITM}=-\frac{1}{B}\sum_{i=1}^B(y_i&\log(p_i)+(1-y_i)\log(1-p_i)),
\end{split}
\end{equation}
where $fc(\cdot)$ is a classifier (e.g. a linear layer followed by a softmax operation). However, as the features of different modalities exist in separate embedding spaces, the transformer-based cross-modal encoders and $\mathcal{L}_{ITM}$ may not possess adequate alignment and fusion capabilities. To address this issue, we introduce a novel contrastive image-text loss function ($\mathcal{L}_{CIT}$).

Different from conventional contrastive loss functions that measure the similarity between embeddings using dot products, we adopt two projection heads $proj_{v}(\cdot)$ and $proj_{t}(\cdot)$ to project the embeddings of different modalities into a generalized low-dimensional space. Then, we measure the similarity of the image and caption by dot product between the projection of $v^{(i)}_{CLS}$ and $t^{(i)}_{CLS}$:
\begin{equation}
\begin{split}
S(V_i,X_i) = proj_{v}(v^{(i)}_{CLS})\cdot proj_{t}(t^{(i)}_{CLS}).
\end{split}
\end{equation}

The similarity between the text and visual representations will be higher for matched text-entity pairs since they express similar content and produce similar representations.
Thus, given a batch of matched image-caption pairs $D_m=\{(X_i, V_i,y_i)\}_{i=1}^{N_m}$, with $N_m<B$ being the size of matched image-caption pairs from a batch and $y_i=1$, we minimize the contrastive image-text loss $\mathcal{L}_{CIT}$ to strengthen the model's ability to align the text-image pairs:

\begin{equation}
\begin{split}
\mathcal{L}_{IT}=-\frac{1}{N_m} &\sum_{i=1}^{N_m} \log \frac{\exp \left(S(V_i,X_i) / \tau\right)}{\sum_{j=1}^{N_m} \exp \left(S(V_i,X_j / \tau\right)},\\
\mathcal{L}_{TI}=-\frac{1}{N_m} &\sum_{i=1}^{N_m} \log \frac{\exp \left(S(V_i,X_i) / \tau\right)}{\sum_{j=1}^{N_m} \exp \left(S(V_j,X_i / \tau\right)},\\
\mathcal{L}_{CIT}&=\frac{1}{2}(\mathcal{L}_{IT}+\mathcal{L}_{TI}),
\end{split}
\end{equation}  
where $\tau$ is the temperature coefficient. The contrastive text-to-image loss $\mathcal{L}_{TI}$ encourages the text embedding to be similar to the image embedding, while the image-to-text loss $\mathcal{L}_{IT}$ is the opposite, aiming to align the image to text embeddings. 

Finally, we define the contrastive image-text loss $\mathcal{L}_{CIT}$ as the average of the text-to-image loss $\mathcal{L}_{TI}$ and the image-to-text loss $\mathcal{L}_{IT}$, which encourages bidirectional alignment of the text and image embeddings, leading to improved multimodal representations.

\subsection{Prompt Guided Entity-Object Alignment}
After establishing the model's image and text alignment capabilities using $\mathcal{L}_{ITM}$ and $\mathcal{L}_{CIT}$, a more refined object-entity alignment ability is necessary to support the MNER task. To this end, the module is designed to generate soft entity pseudo-labels for fine-grained image regions, providing supervision to the pre-training model and improving its object-entity alignment capabilities. Soft pseudo-labels differ from hard pseudo-labels in that they are represented as a probability distribution over types. By generating soft entity pseudo-labels for fine-grained image regions, the model can utilize object information in the image to aid in entity recognition and also improve the model's understanding of the entities that appear in the pre-training dataset. As the encoders defined in section \ref{section:pre-training model}, the soft entity pseudo-label generator consists of a textual encoder and a visual encoder to encode the textual entity prompts and visual objects obtained from the original image, separately.

In order to obtain entities from the dataset to increase the ability of model entity recognition, we first segment the caption of the dataset using spaCy and extract the most frequent M nouns as the candidate entity set $ENTITY$. Next, we generate a series of prompts $PROMPT_e=\{PROMPT_e^{(1)}, PROMPT_e^{(2)},\cdots, PROMPT_e^{(M)}\}$ based on the M candidate entities using a fixed template, for example: ``This is an image of $\{entity\}$'', where $entity\in ENTITY$. We obtain the text representation of these prompts $\{\boldsymbol{t}^{(1)}_{CLS},\boldsymbol{t}^{(2)}_{CLS},\cdots,\boldsymbol{t}^{(M)}_{CLS}\}$ by sending them to the entity text encoder. Subsequently, we use YOLO to crop the input image $V_i$ to generate an object $O_i$ of the image, we then partition $O_i$ into K disjoint small patches and send them to the entity visual encoder to obtain its visual representation $\boldsymbol{o}^{(i)}_{CLS}$. After obtaining the object visual representation and text representations of the prompts, we use the normalized softmax score between the visual representation and all textual representations to generate soft entity pseudo-labels $\boldsymbol{q}^{(i)}_{ent}$:
\begin{equation}
\begin{split}
    \boldsymbol{q}^{(i)}_{ent}=\frac{\exp \left(S\left(O_i, PROMPT_e^{(i)}\right) / \tau\right)}{\sum_{i=1}^M \exp \left(S\left(O_i, PROMPT_e^{(i)}\right) / \tau\right)},
\end{split}
\end{equation} where $\boldsymbol{q}^{(i)}_{ent}\in\mathbb{R}^M$ and $\tau$ is the temperature coefficient.

The soft entity pseudo-label will guide the pre-training by providing supervision for the model's object-entity alignment capabilities. Specifically, let $P^{(i)}_o$ be the subset of $P_i$ which consists of image patches containing the object $O_i$. We use the average of the corresponding multimodal embedding in $P^{(i)}_o$ and send it to a classifier $fc_{ent}(\cdot)$, which outputs probability distribution $P^{(i)}_{ENT}$ among the candidate entities:
\begin{equation}
\begin{split}
    \boldsymbol{e}^{(i)}_{obj} = AVG(&\boldsymbol{e}^{(i)}_{j}), p^{(i)}_j \in P^{(i)}_o,\\
    \boldsymbol{p}^{(i)}_{ent} = &fc_{ent}(\boldsymbol{e}^{(i)}_{obj}),
\end{split}
\end{equation}
where $\boldsymbol{e}^{(i)}_{obj} \in \boldsymbol{R}^d$ and $\boldsymbol{p}^{(i)}_{ent}\in \mathbb{R}^M$. 

Ultimately, we define the contrastive object-entity loss $\mathcal{L}_{COE}$ as the cross-entropy loss between $\boldsymbol{p}_{ent}$ and $\boldsymbol{q}_{ent}$:
\begin{equation}
\begin{split}
    \mathcal{L}_{COE}=-\frac{1}{B}\sum_{i=1}^B\sum_{j=1}^{M} \boldsymbol{q}^{(i)}_{ent,j} \cdot \log \boldsymbol{p}^{(i)}_{ent,j},
\end{split}
\end{equation}where B is the batch size.

\subsection{Prompt Guided Image-Relation Alignment}
After obtaining the alignment ability of entity-object, which helps the model to better discover entities, we need to obtain the alignment ability of relation-image to enhance the model's ability to recognize the relation between entities so that improve the model's performance on MRE tasks that require understanding the relations between entities in text.

This module also contains a textual encoder as well as a visual encoder. We propose generating soft relation pseudo-labels that improve the model's relation-image alignment using a series of prompts based on pre-defined relation tags $RELATION=\{relation_i\}_{i=1}^{L} $. The relational tags are obtained from Wikidata, with a focus on relations of type ``data'' which are the primary relations associated with entities. The use of these relation tags enhances the model's ability to generalize and accurately identify relations between entities. This is similar to our approach for generating soft entity pseudo-labels, which provided supervision for object-entity alignment capabilities. Specifically, we generate a series of prompts $PROMPT_r=\{PROMPT_r^{(1)}, PROMPT_r^{(2)}, PROMPT_r^{(L)}\}$ based on pre-defined relation tags $RELATION$, for example, "The image shows the relation of $\{relation\}$", where $relation \in RELATION$. 

As the image is randomly cropped to obtain the object, it is not guaranteed that the entity mentioned in the sentence corresponds to the object that is cropped from the image. Therefore, we use the entire image as the target for relational alignment.
 We define the soft relation pseudo-labels $\boldsymbol{q}_{rel}$ and the probability distribution $\boldsymbol{p}_{rel}$ obtained from the classifier $fc_{rel}(\cdot)$ as:
\begin{equation}
\begin{split}
    \boldsymbol{q}^{(i)}_{REL}=&\frac{\exp \left(S\left(V, PROMPT_r^{(i)}\right) / \tau\right)}{\sum_{i=1}^{N_r} \exp \left(S\left(V, PROMPT_r^{(i)}\right) / \tau\right)},\\
    &\boldsymbol{p}^{(i)}_{REL} = fc_{rel}(\boldsymbol{e}^{(i)}_{CLS}),
\end{split}
\end{equation}
where we utilize the embedding of the special token \textit{[CLS]} $\boldsymbol{e}_{CLS}$ as the multimodal representation.

Finally, the contrastive image-relation loss $\mathcal{L}_{CIR}$ is defined as:
\begin{equation}
\begin{split}
    \mathcal{L}_{CIR}=-\frac{1}{B}\sum_{i=1}^{B}\sum_{j=1}^{N_r} \boldsymbol{q}^{(i)}_{REL,j} \cdot \log \boldsymbol{p}^{(i)}_{REL,j}.
\end{split}
\end{equation}

Overall, the model is optimized through the joint optimization of four losses, which include: (1) Image-text matching loss ($\mathcal{L}_{ITM}$) and (2) Contrastive image-text loss ($\mathcal{L}_{CIT}$), both of which improve the model's ability to align image and text. (3) Contrastive object-entity loss ($\mathcal{L}_{COE}$), which enhances the model's ability to align objects with corresponding entities. (4) Contrastive image-relation loss ($\mathcal{L}_{CIR}$), which improves the model's ability to align entities with their corresponding relations. Our final loss function is:
\begin{equation}
\mathcal{L}=\lambda_{ITM}\mathcal{L}_{ITM}+\lambda_{CIT}\mathcal{L}_{CIT}+\lambda_{COE}\mathcal{L}_{COE}+\lambda_{CIR}\mathcal{L}_{CIR},
\end{equation}
where the scalar hyper-parameters $\lambda_{ITM}$, $\lambda_{CIT}, \lambda_{COE}$, and $\lambda_{COE}$ are used to control the weight of the four losses.

\begin{table*}
\centering
\caption{Results of different methods on MNER and MRE datasets. $^{\dagger}$ means the results we get directly from \citet{chen2022good}.}
\vspace{-3mm}
\scalebox{0.84}{
\begin{tabular}{cc|ccccccccc}
\toprule
\multicolumn{2}{c|}{\multirow{2}{*}{Methods}} & \multicolumn{3}{c}{Twitter-2015} & \multicolumn{3}{c}{Twitter-2017} & \multicolumn{3}{c}{MRE}  \\ \cmidrule(lr){3-5} \cmidrule(lr){6-8} \cmidrule(lr){9-11} 
& &  Precision & Recall & F1 &  Precision & Recall & F1 &  Precision & Recall & F1  \\
\midrule
\multirow{3}{*}{\makecell[c]{Text \\ -Based}} & \multicolumn{1}{|l|}{CNN-BiLSTM-CRF$^{\dagger}$} & 66.24 & 68.09 & 67.15 & 80.00 & 78.76   & 79.37 &-- &-- &-- \\  

& \multicolumn{1}{|l|}{BERT-CRF$^{\dagger}$} & 69.22  & 74.59 & 71.81 & 83.32 & 83.57   & 83.44 &-- &-- &-- \\

& \multicolumn{1}{|l|}{MTB$^{\dagger}$} &  & -- & -- & -- &--   &-- &64.46 &57.81 &60.86 \\

\midrule
\midrule

\multirow{10}{*}{\makecell[c]{Multimodal \\ -Based}} & \multicolumn{1}{|l|}{AdapCoAtt-BERT-CRF$^{\dagger}$} & 69.87 & 74.59 & 72.15 & 85.13 & 83.20   & 84.10 &--  &-- &-- \\  

& \multicolumn{1}{|l|}{OCSGA$^{\dagger}$} & 74.71  & 71.21 &72.92 & -- &--   &-- &--  &-- &-- \\  

& \multicolumn{1}{|l|}{UMT$^{\dagger}$} & 71.67 & 75.23 & 73.41 & 85.28 &85.34   &85.31 &62.93  &63.88 &63.46 \\ 

& \multicolumn{1}{|l|}{UMGF$^{\dagger}$} & 74.49 & 75.21 & 74.85 &86.54 &84.50   &85.51 &64.38  &66.23 &65.29 \\ 

& \multicolumn{1}{|l|}{BERT+SG$^{\dagger}$} &--  &-- &-- &-- &-- & --  &62.95  &62.65 &62.80 \\ 

& \multicolumn{1}{|l|}{MEGA$^{\dagger}$} & 70.35  & 74.58 & 72.35  &84.03 &84.75   & 84.39 & 64.51  &68.44 &66.41 \\ 

& \multicolumn{1}{|l|}{IFAformer} & -- & -- & -- & -- &--   &-- &82.59  &80.78 &81.67 \\ 

& \multicolumn{1}{|l|}{HVPNeT$^{\dagger}$} & 73.87 & 76.82 & 75.32 & 85.84 &87.93   &86.87 &83.64  &80.78 &81.85 \\ 

& \multicolumn{1}{|l|}{MoRe}   & 
 77.62\small±0.63 & \underline{80.52\small±1.22} &\underline{79.04\small±1.04} &\underline{90.35\small±0.79}   &\underline{90.41\small±0.46} &\underline{90.38\small±0.63}  &66.66\small±1.35 &70.58\small±1.07 &68.56\small±1.14\\ 

& \multicolumn{1}{|l|}{MMIB} & 74.38\small±1.04 & 77.72\small±1.21 & 
 76.01\small±1.13 & 87.42\small±0.93 &87.68\small±0.63   & 87.55\small±0.78 &\underline{83.49\small±1.14}  &\underline{82.97\small±0.68} &\underline{83.23\small±0.89} \\ 

\midrule
\midrule

\multirow{4}{*}{\makecell[c]{Multimodal \\ Pre-training}} & \multicolumn{1}{|l|}{CLIP} & 74.25\small±1.13 & 74.64\small±1.05 & 74.44\small±1.08 & 85.34\small±0.94 &85.29\small±1.20   &85.31\small±1.08 &78.43\small±0.91 &77.39\small±0.99 &77.91\small±0.95 \\  

& \multicolumn{1}{|l|}{Oscar} & 74.02\small±1.10 & 74.15\small±0.86 & 74.08\small±0.96 & 84.76\small±0.92 &84.82\small±1.21   &84.79\small±1.06 &78.59\small±0.82 &77.51\small±0.79 &78.05\small±0.81 \\  

& \multicolumn{1}{|l|}{U-VisualBERT} & 76.28\small±1.08 & 75.41\small±1.13 & 75.84\small±1.11 & 87.43\small±0.96 &86.25\small±0.89   &86.84\small±0.93 &80.24\small±1.27 &79.15\small±0.93 &79.69\small±1.11 \\ 

& \multicolumn{1}{|l|}{BLIP} & \underline{77.73\small±1.07} & 76.58\small±0.86 & 77.15\small±0.97 & 88.92\small±1.13 &88.67\small±1.02   &88.79\small±1.06 &82.42\small±0.86 &81.98\small±0.98 &82.20\small±0.92 \\ 

\midrule
\midrule

\multicolumn{2}{c|}{Ours}  &\textbf{80.03\small±0.79} & \textbf{80.97\small±0.68}  & \textbf{80.50\small±0.82} & \textbf{91.97\small±0.85} & \textbf{91.33\small±1.04} & \textbf{91.65\small±0.79}  & \textbf{84.95\small±0.82}  & \textbf{85.76\small±0.77}& \textbf{84.86\small±0.85}\\

\multicolumn{2}{c|}{Ours \textit{w/o Entity-Object Alignment}} & 76.94\small±1.17 & 77.28\small±1.26  & 77.11\small±1.21 & 87.93\small±0.97 &87.52\small±1.02 & 87.67\small±1.00  &  83.24\small±1.09 &  83.17\small±0.97 &83.20\small±1.03 \\

\multicolumn{2}{c|}{Ours \textit{w/o Relation-Image Alignment}} & 77.83\small±1.18 & 78.52\small±1.02  & 78.17\small±1.10 & 88.74\small±1.07 &88.28\small±1.01 & 88.51\small±1.04  &  82.95\small±1.10 &  82.88\small±0.99 &  82.91\small±1.05  \\

\multicolumn{2}{c|}{Ours \textit{w/o Soft Pseudo Labels}} & 78.34\small±1.02 & 78.97\small±1.09  & 78.65\small±1.06 & 89.82\small±0.93 &89.45\small±0.88 & 89.63\small±0.90  &  84.34\small±1.06 &  84.06\small±1.00 &  84.20\small±1.03  \\

\bottomrule
\end{tabular}}
\label{tab:verification}
\vspace{-4mm}
\end{table*}

\subsection{Model Fine-tuning on MNER and MRE}
After pre-training the model, we fine-tune it in MNER \& MRE tasks. 
\subsubsection{MNER}
For a batch of multimodal MNER data $\mathcal{X}_{MNER}=\{X_i,V_i,Y_i\}_{i=1}^{B}$ defined in Sec. \ref{sec:task}, where $B$ represents the batch size, we input the textual and visual data into the pre-trained model and utilize the output $e_{MNER}=\{e_1,e_2,\cdots,e_N\}$ of the multimodal encoder to obtain a representation for each position in the sentence.

Subsequently, we pass $e_{MNER}$ to a conditional random field (CRF) to enforce structural correlations between labels in sequence decoding. For a specific forecast sequence $Y_{pre}=\{y_{pre}^1,y_{pre}^2,\cdots,y_{pre}^N\}$, the probability of the sequence is defined as follows:
\begin{equation}
\begin{split}
p\left(Y_{pre} | e_{MNER}\right)=\frac{\prod_{i=1}^N S_i\left(Y_{pre}^{i-1}, Y_{pre}^i, e_{MNER}\right)}{\sum_{y^{\prime} \in \mathcal{Y}} \prod_{i=1}^N S_i\left(Y_{pre}^{'i-1}, Y_{pre}^{'i}, e_{MNER}\right)},
\end{split}
\end{equation}
where $\mathcal{Y}$ is the pre-defined label set with \texttt{BIO} tagging schema, and $S(\cdot)$ represents the potential function. Finally, the output label sequence $Y^{*}$ and the training loss are:
\begin{equation}
\begin{split}
Y^* &= \arg\max_{Y_{pre}} p(Y_{pre}|e_{MNER}),\\
\mathcal{L}_{MNER} &= -\frac{1}{B}\sum_{i=1}^B\sum_{j=1}^{N}\log(p(Y_i^j|e_{MNER})).
\end{split}
\end{equation}
\subsubsection{MRE}
Given a batch of multimodal MRE data, denoted as $\mathcal{X}_{MRE}=\{X_i, V_i, y_i\}_{i=1}^B$, where $B$ is the batch size. To process the sentence $X=\{x_1, x_2,\cdots, x_N\}$ along with its corresponding entities $E1$ and $E2$, we follow the approach proposed by ~\citet{soares2019matching} by introducing four special tokens to represent the start and end of $E1$ and $E2$. These tokens are denoted as $[E1_{start}]$, $[E1_{end}]$, $[E2_{start}]$, and $[E2_{end}]$, and are injected into the original sentence $X$:
\begin{equation}\label{embedding}
\begin{split}
       {X}=&[x_1,\cdots,[E1_{start}],x_i,\cdots,x_{j-1},[E1_{end}],\\ &\cdots ,[E2_{start}],x_k,\cdots,x_{l-1},[/E2_{end}...],\cdots,x_N].
\end{split}
\end{equation} We then utilize this modified token sequence as the input for the textual encoder. For the relation representation between entities $E1$ and $E2$, we concatenate contextualized entity representations corresponding to $[E1_{start}]$ and $[E2_{start}]$ tokens from the multimodal encoder, obtaining $\boldsymbol{e}_{rel}\in \mathbb{R}^{2\times d}$, where $d$ is the hidden dimension of 768.

After obtaining the fixed-length relation representation $\boldsymbol{e}_{rel}$, we pass it through the classifier $fc_{MRE}(\cdot)$ to obtain the probability distribution $p(y | X)=fc(\boldsymbol{e}_{rel})$ over the different classes. The pre-trained model is fine-tuned using the cross-entropy loss function:
\begin{equation}
    \mathcal{L}_{MRE}=\frac{1}{B}\sum_{i=1}^{B}loss(y_i,p(y|x_i)),
\end{equation}

\section{Experiments and Analyses}
\label{sec:experiments}

\begin{table}
\centering
\caption{The F1 performance obtained by using multimodal pre-training on state-of-the-art multimodal method.}
\vspace{-3mm}
\scalebox{0.87}{
\begin{tabular}{c|ccc}
\toprule
\multicolumn{1}{c|}{Methods} & \multicolumn{1}{c}{Twitter-2015} & \multicolumn{1}{c}{Twitter-2017} & \multicolumn{1}{c}{MRE}  \\ \midrule
MMIB &  76.01 & 87.55 & 83.23  \\  
\midrule
MMIB+CLIP & 79.23 & 89.69 & 85.05 \\  
MMIB+Oscar & 80.53 & 90.47 & 85.87  \\  
MMIB+U-VisualBERT &  80.72 & 90.26 & 85.49  \\  
MMIB+BLIP &  82.06 & 91.69 & 86.34  \\  
MMIB+Ours &  \textbf{82.94} & \textbf{92.79} & \textbf{87.46}  \\  

\bottomrule
\end{tabular}}
\label{tab:continual}
\vspace{-5mm}
\end{table}
\subsection{Experimental Setup}
\textbf{Dataset}: In line with prior research \cite{chen2022good,cui2023enhancing}, we carry out experiments on Twitter-2015 \cite{zhang2018adaptive} and Twitter-2017 \cite{lu2018visual} datasets containing 4,000/1,000/3,257, 3,373/723/723 sentences in train/dev/test sets for MNER task, and the Multimodal Relation Extraction (MRE) dataset \cite{zheng2021mnre} containing 12,247/1,624/1,614 sentences in train/dev/test sets. These datasets consist of multimodal Twitter posts, with each post featuring an image and a text snippet. Both Twitter-2015 and Twitter-2017 include four entity categories: Person (PER), Location (LOC), Organization (ORG), and Miscellaneous (MISC). Meanwhile, the MRE dataset comprises 23 relation types. 

\noindent \textbf{Evaluation Metrics}: For MNER, an entity is deemed accurately identified if both its span and entity type correspond with the gold standard answer. In the case of MRE, a correct extraction of the relation between two entities occurs when the predicted relation type aligns with the gold standard. We utilize the evaluation code provided by \citet{chen2022good}, which employs overall Precision, Recall, and F1-score for assessment purposes.

\noindent \textbf{Hyperparameters}:
In our study, we employed $BERT_{base}$ as the text encoder, setting the maximum token length to 80. To serve as the visual encoder, we utilized a visual transformer. Our optimization strategy involved the AdamW optimizer with a decay rate of 1e-3 and a learning rate of 1e-4. In addition, the scale hyperparameters were configured as follows: $\lambda_{ITM}=1$, $\lambda_{CIT}=1$, $\lambda_{COE}=1$, and $\lambda_{CIR}=1$.
We pre-trained our model on NewsCLIPping using 8 NVIDIA 3090 GPUs with a batch size of 128 for 85,000 iterations and then fine-tuned the model using a batch size of 16.

\vspace{-2mm}
\subsection{Baselines}
Following previous approaches \cite{chen2022good}, we concentrate on three categories of models for this comparison: text-based models, earlier MNER and MRE models, and multimodal pre-training methods.

Text-based models: 1) CNN-BiLSTM-CRF \cite{ma2016end}; and 2) BERT-CRF for NER \cite{devlin2019bert}. The subsequent models are designed specifically for RE: 3) MTB \cite{soares2019matching}, an RE-focused pretraining model built on BERT.

Prior MNER and MRE models: 1) AdapCoAtt-BERT-CRF \cite{zhang2018adaptive}; 2) OCSGA \cite{wu2020multimodal}; 3) UMT \cite{yu2020improving}; 4) UMGF \cite{zhang2021multi} introduces a unified multi-modal graph fusion method for MNER. 5) BERT+SG \cite{zheng2021multimodal}, merges textual representations from BERT with visual features produced using a scene graph (SG) tool \cite{tang2020unbiased}. 6) MEGA \cite{zheng2021multimodal}, formulates a dual graph for multi-modal alignment to better capture the correlation between entities and objects. 7) MoRe \cite{wang2022named}, uses a retrieval approach to search for a large number of related documents on Wikipedia as auxiliary representations for both images and text. 8) IFAformer \cite{li2022analyzing}. 9) HVPNeT \cite{chen2022good}, enhances hierarchical detection in image object recognition, removing objects unrelated to entities. 10) MMIB \cite{cui2023enhancing}, introduces Information Bottleneck to remove noise in different modalities and aligns multimodal data.

Multimodal pre-training methods: 1) CLIP \cite{radford2021learning}, 2) BLIP \cite{li2022blip}, 3) Oscar \cite{li2020oscar}, and 4) U-VisualBERT \cite{li2021unsupervised} all use text to align images or objects. In addition, our pre-training method is orthogonal to previous MNER and MRE models, so we also add our pre-training experiments to prior SOTA model to compare whether its performance will further increase.

\subsection{Main Results}

Table \ref{tab:verification} shows the results of our multimodal pre-training approach and three baseline methods on three public datasets, with each test conducted thrice. From comparing these methods, we can conclude: \textbf{(1)} Multimodal-based and pre-training methods outperform text-based entity and relation extraction techniques, due to the integration of image information in multimodal training. \textbf{(2)} Our pre-training method improves the F1 score by an average of 3.41\% over the previous best multimodal method (MMIB) on the three datasets, demonstrating that adding numerous unlabeled object-entity and image-relation alignments during pre-training greatly enhances the model's ability to utilize visual object information and leverage object interactions for relation classification in text. \textbf{(3)} Interestingly, the baseline model MoRe achieves the previous best result on the multimodal NER dataset but performs poorly on the multimodal RE dataset (14.67\% lower F1 than MMIB) because it loses object input during multimodal fusion, leading to a lack of essential object modality information during entity relation extraction. \textbf{(4)} As all multimodal pre-training methods are orthogonal to multimodal fusion methods, we selected the best fusion method (MMIB) and continued using multimodal pre-training based on MMIB. Results in Table \ref{tab:continual} show that all pre-training methods can further enhance MMIB's F1 performance, with ours achieving a new record, increasing the initial performance by 6.93\%, 5.24\%, and 4.23\% on the three datasets, respectively.

\begin{figure}[t!]
\centering
\includegraphics[width=0.84\linewidth]{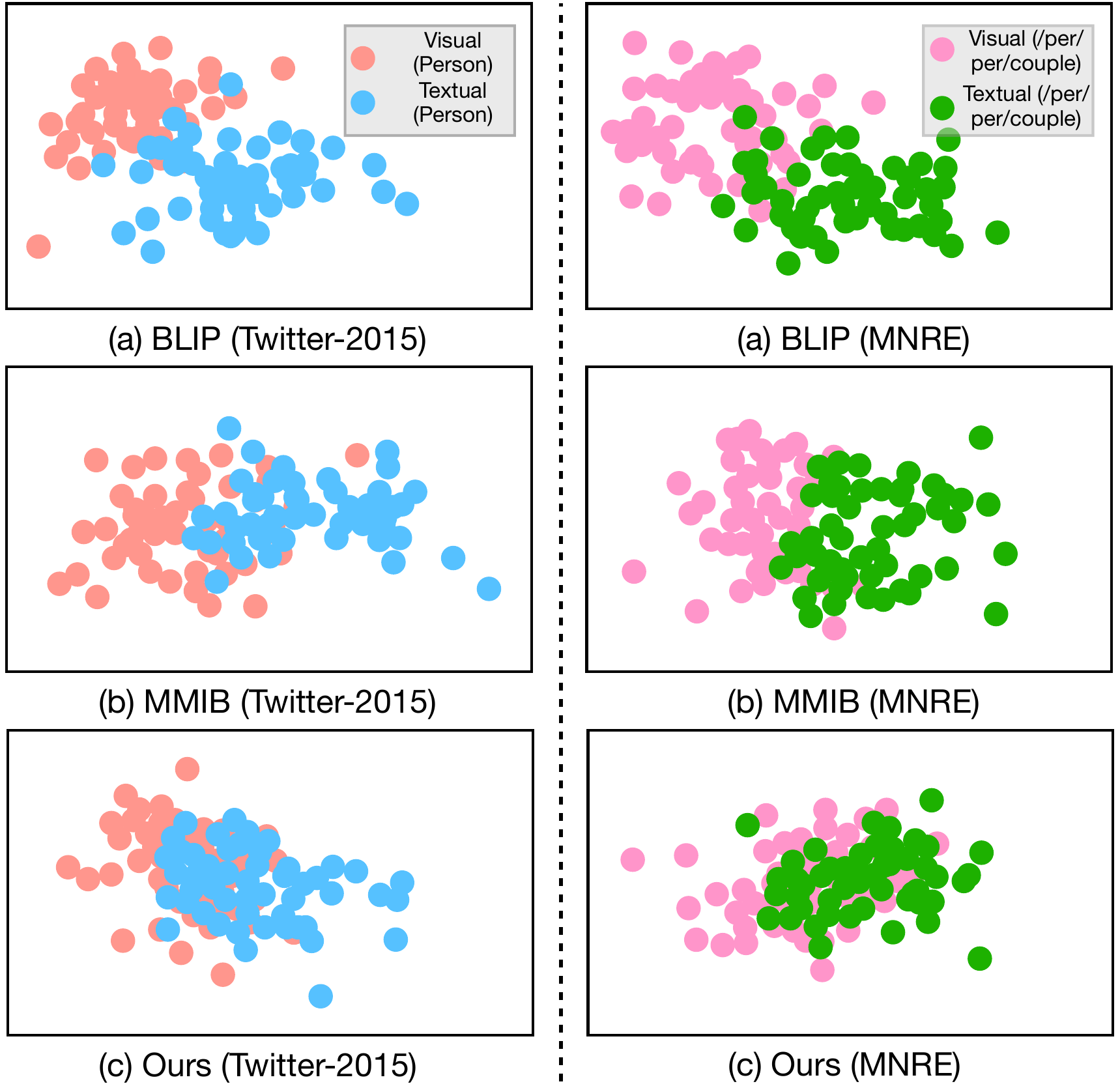}
\vspace{-2mm}
\caption{Consistency of Entity-Object and Image-Relation Representations in Semantic Space.}
\label{fig:consistency}
\vspace{-5mm}
\end{figure}

\begin{table}
\centering
\caption{The F1 Impact of Combining Prompt Designs.}
\vspace{-3mm}
\scalebox{0.99}{
\begin{tabular}{c|cccccc}
\toprule
\multicolumn{1}{c|}{Datasets} & \multicolumn{1}{c}{{\Large \ding{172}}+$\mathbb{A}$} & \multicolumn{1}{c}{{\Large \ding{172}}+$\mathbb{B}$} & \multicolumn{1}{c}{{\Large \ding{172}}+$\mathbb{C}$} & \multicolumn{1}{c}{{\Large \ding{173}}+$\mathbb{A}$}& \multicolumn{1}{c}{{\Large \ding{173}}+$\mathbb{B}$}& \multicolumn{1}{c}{{\Large \ding{173}}+$\mathbb{C}$} \\ \midrule
Twitter-2015 &  \textbf{80.50} &  80.18 & 80.49  & 80.32 & 79.97 & 80.14  \\  
Twitter-2017 & \textbf{91.65} & 90.77 &  91.21  &  90.90 &  90.58 &  91.02 \\  
MRE & 84.86 & 84.97 &  \textbf{85.02} &  84.53 & 84.01 & 84.47\\

\bottomrule
\end{tabular}}
\label{tab:prompt}
\vspace{-4mm}
\end{table}

\begin{figure}[t!]
\centering
\includegraphics[width=0.99\linewidth]{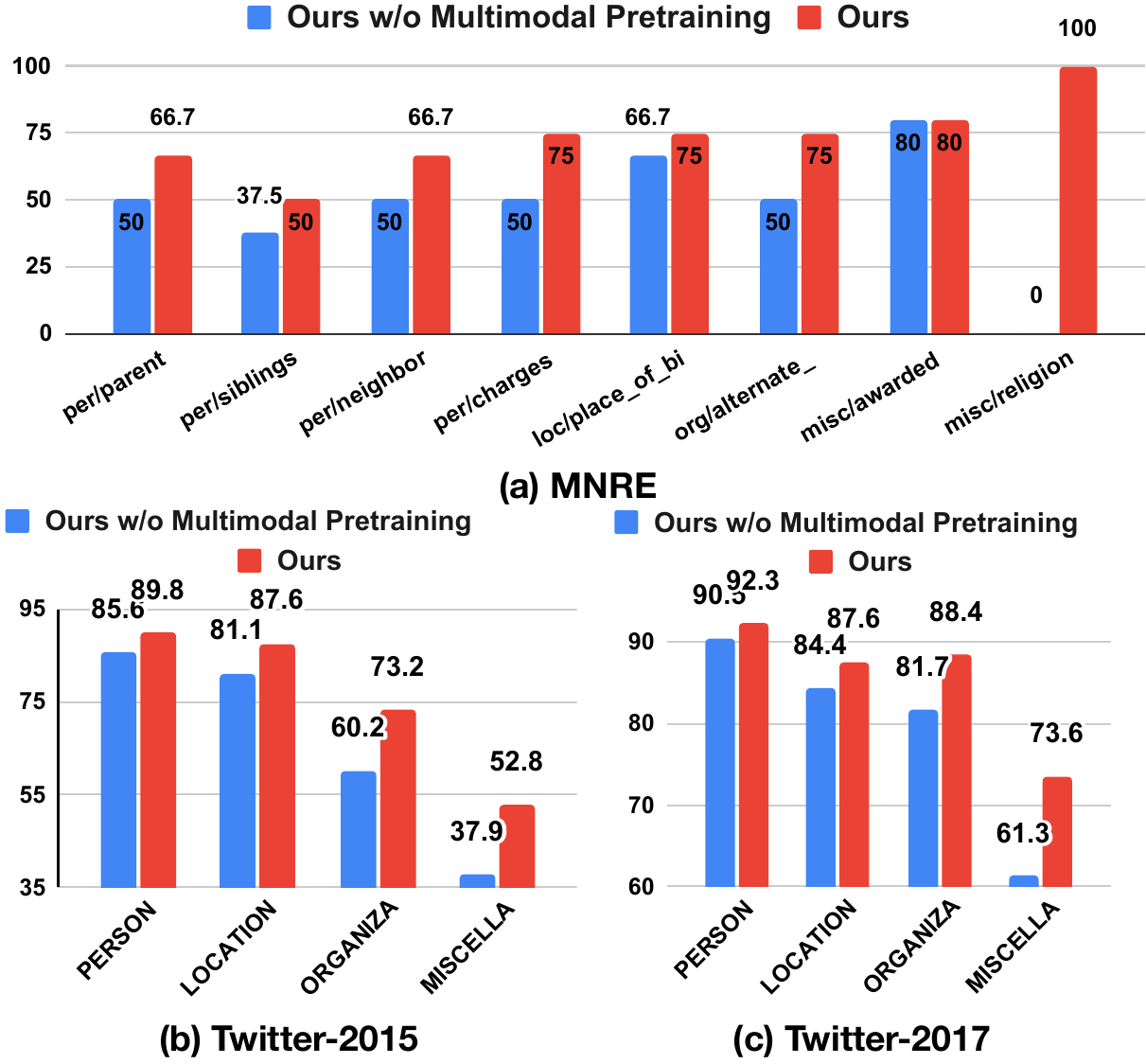}
\vspace{-2mm}
\caption{The F1 Scores Improvement in Tail Relations and Entity Types.}
\label{fig:tail_change}
\vspace{-6mm}
\end{figure}

\subsection{Analyses and Discussions}

\paragraph{Ablation Study}
We conducted ablation experiments on three modules in the multimodal pre-training tasks to demonstrate their effectiveness. Ours \textit{w/o Entity-Object Alignment} and Ours \textit{w/o Relation-Image Alignment} represent the removal of entity-object alignment and relation-image alignment tasks during pre-training, which results in a reduced degree of learning for entities or relations by the multimodal fusion module. From Table \ref{tab:verification}, entity-object alignment and relation-image alignment tasks contribute an average F1 improvement of 3.01\% and 2.47\%, respectively, for the multimodal fusion module. Since Entity-Object Alignment benefits both NER and RE tasks, it has a more significant impact. Ours \textit{w/o Soft Pseudo Labels} uses hard pseudo-labels instead, reducing the fusion module's error-smoothing ability; incorrect predictions directly affect the module, causing an average F1 drop of 1.51\%.

\begin{figure}[t!]
\centering
\includegraphics[width=0.99\linewidth]{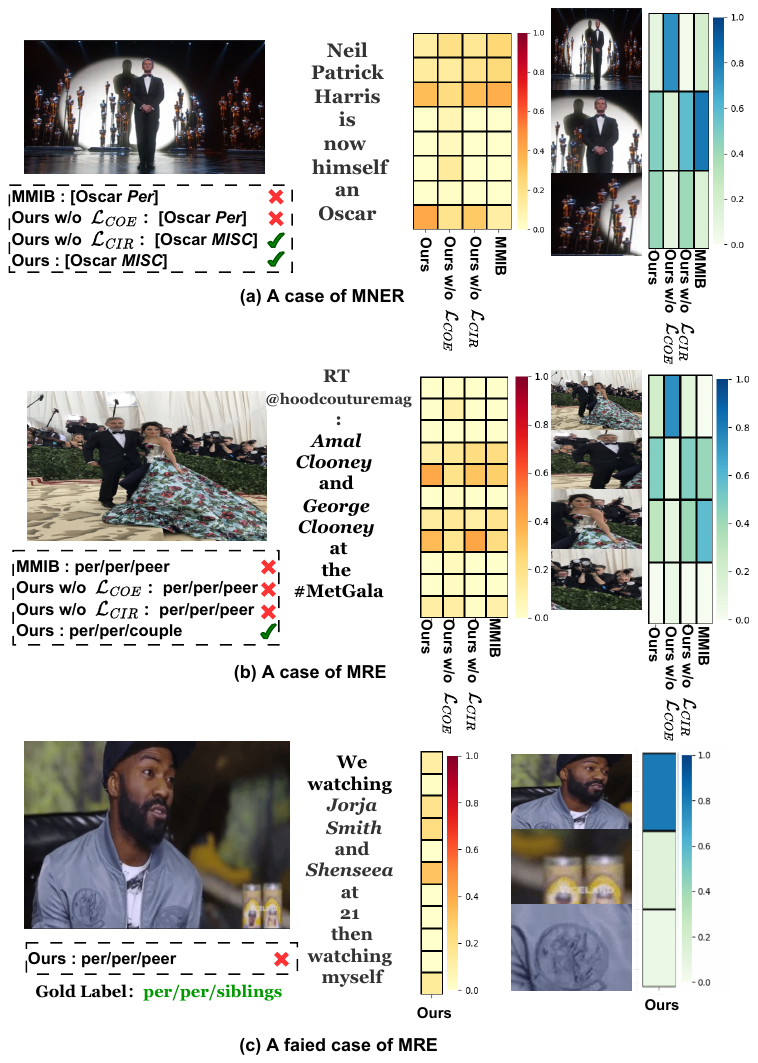}
\vspace{-2mm}
\caption{Case Study and Error Analysis of MNER and MRE.}
\label{fig:case}
\vspace{-6mm}
\end{figure}

\paragraph{Investigating the Consistency of Entity-Object and Image-Relation Representations in Semantic Space}

We displayed entity-object and image-relation representations in a semantic space to show if different modalities' information is well combined. We chose 50 test samples from Twitter-2015 and MRE datasets and used YOLO \cite{wang2022yolov7} to get objects. We applied t-SNE \cite{hinton2002stochastic} to reduce the text and image embeddings to 2D after modality fusion and plotted them in Figure \ref{fig:consistency}. Our method aligns entity-object and image-relation semantics during pre-training using prompts, significantly helping to align data from various modalities in semantic space and better utilizing multimodal data for entity and relation extraction. Although MMIB and BLIP used entity-object alignment training, the alignment of entity-object and image-relation in semantic space remains sparse due to the absence of prompt templates as a pre-training method.

\paragraph{The Impact of Prompt Design}
We designed various prompts to investigate their influence on pre-trained models. For entity-object-aligned prompts, we used two types: ({\Large \ding{172}}) This is an image of [ENTITY]. ({\Large \ding{173}}) An image of [ENTITY] is shown here. For relation-image-aligned prompts, we employed three types: ($\mathbb{A}$) This image shows the relation of [RELATION]. ($\mathbb{B}$) The relation of [RELATION] is shown in this image. ($\mathbb{C}$) The relation between the objects in the image is [RELATION]. The F1 impact of using different prompts on the results is shown in Table \ref{tab:prompt}. We discovered that prompts in the active voice yield better results than those in the passive voice, though the difference is not significant, with the impact being within 1\%. Of course, our exploration of prompt engineering is limited, which presents an interesting direction for future research.

\paragraph{Investigating the F1 Improvement in Tail Relations and Entity Types}

We examined the F1 improvements in challenging cases of datasets (with fewer examples and greater difficulty) using our multimodal pre-training. Figure \ref{fig:tail_change} shows that for the 8 tail relations, multimodal pre-training impressively boosts the average F1 score by 25.5\%, compared to 9.9\% for all relations. This suggests that multimodal pre-training is more beneficial for RE tasks with less data and lower metrics. Likewise, in multimodal NER tasks, the model achieves an average improvement of 14.0\% and 9.5\% for the less frequent \texttt{ORG} and \texttt{MIS} entity types, significantly surpassing the overall average improvement of 9.7\% and 6.0\% for all entity types. 

\begin{figure}[t!]
\centering
\includegraphics[width=0.99\linewidth]{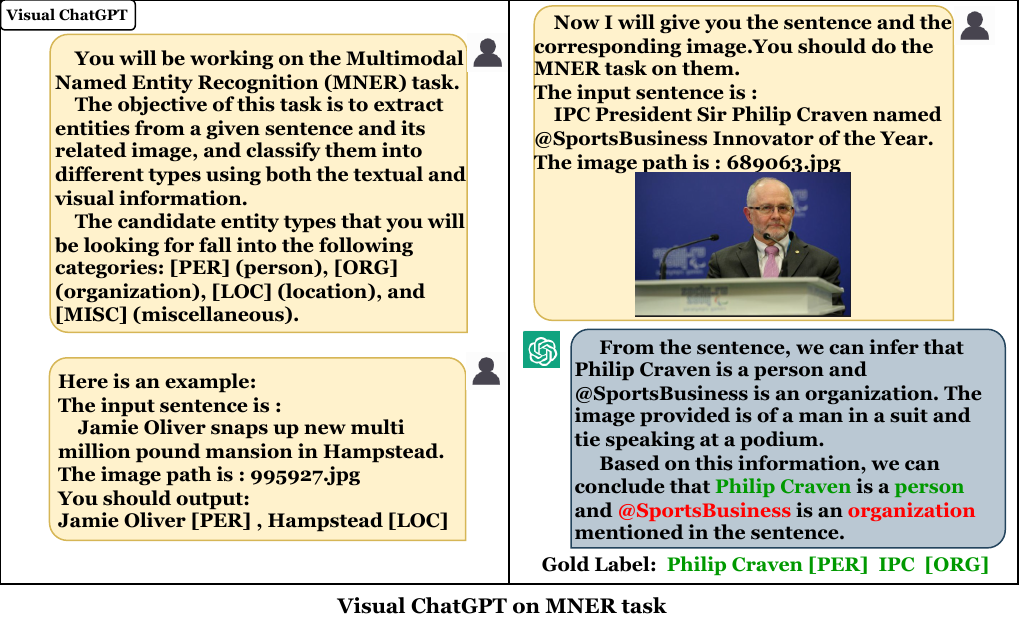}
\vspace{-2mm}
\caption{Study on the Performance of Visual ChatGPT.}
\label{fig:gpt}
\vspace{-6mm}
\end{figure}

\paragraph{Case Study and Error Analysis}

In Fig. \ref{fig:case}, we show a multimodal entity and relation extraction case study, illustrating the efficacy of pre-trained entity-object and relation-image alignment. By standardizing the model's attention to tokens and objects, we can evaluate the model's token and object information usage based on attention strength from 0 to 1. Without entity-object alignment (Ours w/o $\mathcal{L}_{COE}$ and MMIB), the MNER case misaligns ``Oscar'' and misidentifies it as a \texttt{Person}. Our model, with strong prior knowledge, accurately aligns ``Oscar'' with the trophy and identifies it as \texttt{MISC}. In the MRE case, our model and MMIB align ``Amal Clooney'' and ``George Clooney'' with the individuals in the image but only identify the relation as \texttt{per/per/*}. Our model obtains cues like ``The image shows the relation of [spouse]'', ``The image shows the relation of [place of marriage]'', and related image hints, enabling accurate determination of \texttt{per/per/couple}.

Additionally, we present an error analysis where, in the final case, both the image and text fail to communicate the more specific relationship between ``Jorja Smith'' and ``Shenseea''. As a result, our method mistakenly confuses the similar relations: \texttt{per/per/peer} and \texttt{per/per/siblings}, in a less fluent and professional way.

\paragraph{Study on the Performance of ChatGPT}

Recently, ChatGPT has attracted widespread attention for its impressive performance in low-resource scenarios. Although GPT 4.0 also possesses multimodal capabilities, due to the unavailability of its API, we have only used Visual ChatGPT \cite{wu2023visual} for experimenting with MNER and MRE tasks. As shown in Figure \ref{fig:gpt}, we provided two cases as prompt for Visual ChatGPT. For MNER task, Visual ChatGPT incorrectly classified ``SportBusiness'' as an \texttt{ORG}, likely due to its association with the name of a magazine. 
Visual ChatGPT achieved F1 scores of \textbf{41.45}, \textbf{47.24}, and \textbf{36.82} on the test sets of Twitter 2015, Twitter 2015, and MRE, respectively, which still fall short of the SOTA. Of course, further improvements to Visual ChatGPT's performance can be achieved by using more in-context learning cases or employing better prompts, which will be a future research direction.
\section{Conclusion}
\label{conclusion}

In this study, we introduce a new method for multimodal entity and relation extraction using unlabeled image-caption pairs. We propose alignment tasks for entity-object and relation-image pairs using prompts, delivering soft pseudo labels for guiding the pretraining.  This reduces modality misalignment, boosting F1 scores in multimodal NER and RE. Additionally, our self-supervised approach improves hierarchical and graph-based fusion, enhancing SOTA model results.
\begin{acks}
The work was supported by in part by the National Key Research and Development Program of China (No. 2019YFB1704003), the National Nature Science Foundation of China (No. 62021002), NSF under grants III-1763325, III-1909323, III-2106758, SaTC-1930941, Tsinghua BNRist and Beijing Key Laboratory of Industrial Bigdata System and Application. Junzhe Chen is supported by Tsinghua University Initiative Scientific Research Program.
\end{acks}

\balance
\bibliographystyle{ACM-Reference-Format}
\bibliography{sample-base}










\end{document}